\documentclass[final]{cvpr}

\usepackage{times}
\usepackage{epsfig}
\usepackage{graphicx}
\usepackage{amsmath}
\usepackage{amssymb}
\usepackage{booktabs} 
\usepackage{color}
\usepackage{float}
\usepackage{array}
\usepackage{multirow}
\usepackage{amssymb}
\usepackage{caption}
\usepackage{soul}
\usepackage{xcolor}
\usepackage{subfig}
\usepackage[OT1]{fontenc}
\usepackage{bm}

\usepackage{array}
\newcommand{\PreserveBackslash}[1]{\let\temp=\\#1\let\\=\temp}
\newcolumntype{C}[1]{>{\PreserveBackslash\centering}p{#1}}
\newcolumntype{R}[1]{>{\PreserveBackslash\raggedleft}p{#1}}
\newcolumntype{L}[1]{>{\PreserveBackslash\raggedright}p{#1}}

\usepackage[pagebackref=true,breaklinks=true,colorlinks,bookmarks=false]{hyperref}

\def\cvprPaperID{4506} 
\def\confYear{CVPR 2021}
\makeatletter
\def\thanks#1{\protected@xdef\@thanks{\@thanks
        \protect\footnotetext{#1}}}
\makeatother

\begin{document}

\title{Multi-Scale Aligned Distillation for Low-Resolution Detection}

\author{
Lu Qi$^{1\dagger}$\thanks{$\dagger$Equal contribution.},~
Jason Kuen$^{2\dagger}$,~
Jiuxiang Gu$^{2}$,~
Zhe Lin$^2$,~
Yi Wang$^{1}$,~
Yukang Chen$^{1}$,~
Yanwei Li$^1$,~
Jiaya Jia$^{1,3}$
\\[0.2cm]
$^1$The Chinese University of Hong Kong~~
$^2$Adobe Research~~
$^3$SmartMore
}

\maketitle
\thispagestyle{empty}
\pagestyle{empty}

\begin{abstract}
In instance-level detection tasks (e.g., object detection), reducing input resolution is an easy option to improve runtime efficiency. However, this option traditionally hurts the detection performance much.
This paper focuses on boosting performance of low-resolution models by distilling knowledge from a high- or multi-resolution model. We first identify the challenge of applying knowledge distillation (KD) to teacher and student networks that act on different input resolutions. To tackle it, we explore the idea of spatially aligning feature maps between models of varying input resolutions by shifting feature pyramid position and introduce \textbf{aligned multi-scale training} to train a multi-scale teacher that can distill its knowledge to a low-resolution student. Further, we propose \textbf{crossing feature-level fusion} to dynamically fuse teacher's multi-resolution features to guide the student better. On several instance-level detection tasks and datasets, the low-resolution models trained via our approach perform competitively with high-resolution models trained via conventional multi-scale training, while outperforming the latter's low-resolution models by $2.1\%$ to $3.6\%$ in terms of mAP. Our code is made publicly available at \href{https://github.com/dvlab-research/MSAD}{https://github.com/dvlab-research/MSAD}.
\end{abstract}

\section{Introduction}
Deep learning~\cite{krizhevsky2017imagenet,simonyan2014very,szegedy2015going,szegedy2016rethinking,DBLP:conf/cvpr/HeZRS16,szegedy2016inception,huang2017densely,hu2018squeeze} has enabled instance-level detection (object detection~\cite{girshick2014rich,dollar2015fast,DBLP:conf/nips/RenHGS15}, instance segmentation~\cite{he2017mask,DBLP:journals/corr/abs-1803-01534}, human keypoint detection~\cite{simon2017hand,zhou2019objects,sun2019deep}, etc.) methods to achieve previously unattainable performance. Heavy computation requirement of deep-learning-based instance-level detection models, however, remains an issue for easy adoption of these models in real-world applications~\cite{morrison2018cartman, shu2014human, qi2019amodal, chu2019vehicle}.

While many model compression techniques~\cite{polino2018model,chin2020towards,li2019exploiting,kim2018paraphrasing} have been proposed to train compact models for accelerated inference, they mostly focus on trimming networks along depth or width~\cite{ding2019global,zhuang2018discrimination,gao2018dynamic,zhu2017prune,carreira2018learning}, or adopting efficient block structure design~\cite{howard2017mobilenets,sandler2018mobilenetv2,howard2019searching,zhang2018shufflenet,ma2018shufflenet,he2018amc,chen2019detnas}. Besides depth/width, another critical dimension for the compound scaling of network architectures is the input resolution~\cite{tan2019efficientnet,li2020learning}. However, reducing input resolutions to accelerate instance-level detection is generally not regarded as a decent solution in existing work due to severe performance degradation. For example, for the recent one-stage detector FCOS~\cite{tian2019fcos}, its mean average precision (AP) drops from 38.7 to 34.6 when the detector is naively trained on 400px images instead of the default 800px images. 

We are thus interested to study the fundamental problem to {\it upgrade performance of a low-resolution detection model up to that of its high-resolution counterpart}.

\begin{figure}[t!]
	\begin{center}
		\includegraphics[width=\linewidth]{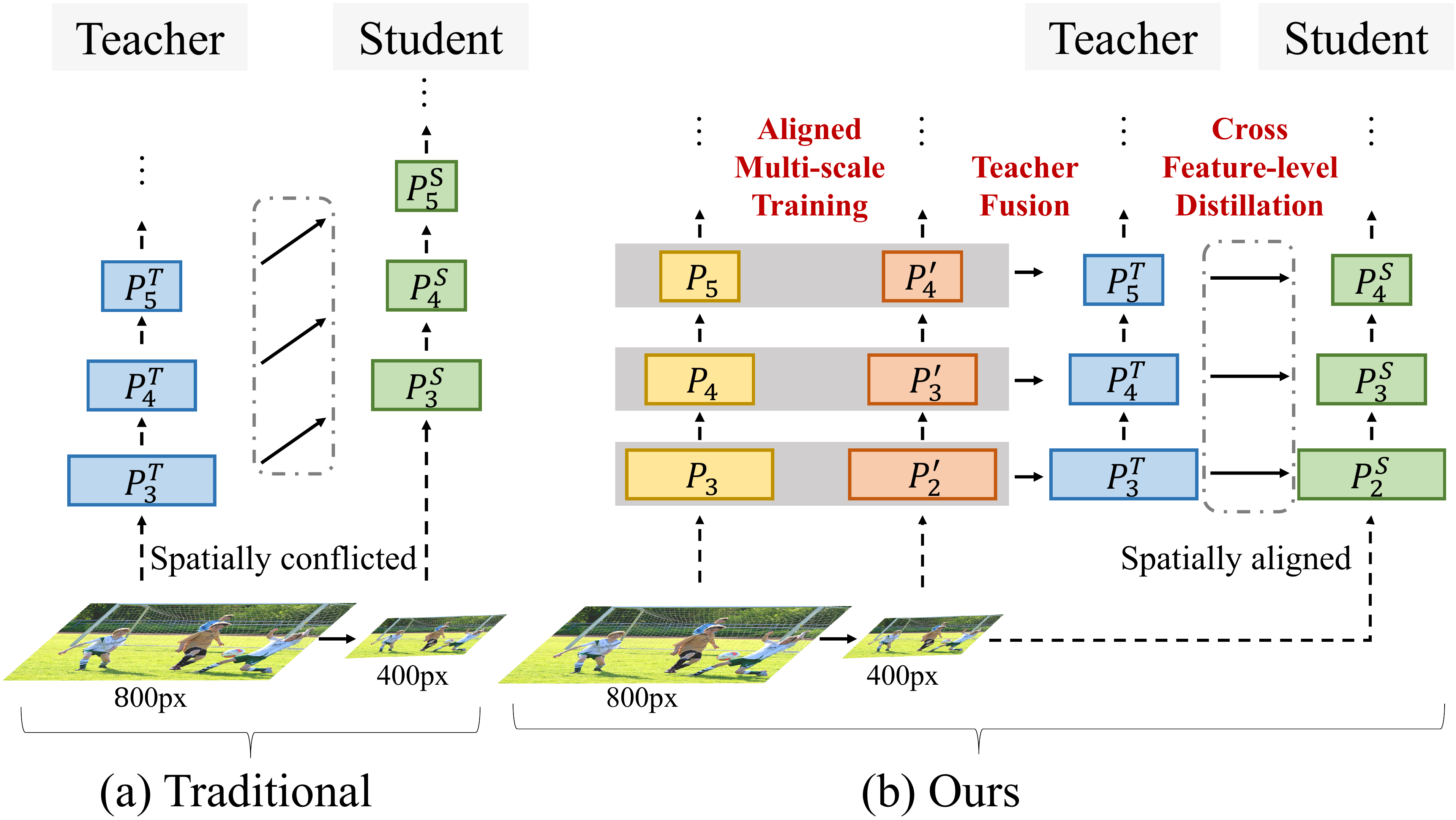}
	\end{center}
	\vspace{-15pt}
	\caption{
	Conceptual comparison between (a) traditional teacher-student approach and (b) ours, in the setting of using a high-resolution teacher to guide a low-resolution student. In this setting, the traditional approach of transferring knowledge along the same feature levels fails due to spatially-conflicted feature maps. To resolve it, we introduce a multi-scale aligned distillation approach.
	} 
	\label{fig:sketch}
\end{figure}

There was study to mitigate performance drop by distilling knowledge (KD) from a high-res teacher to a low-res student \cite{li2020learning,yang2020mutualnet}.
KD methods distill knowledge from a heavier teacher network to a compact student mostly in the context of image classification~\cite{yuan2020revisiting,guo2020online,romero2014fitnets}, since the spatial scales of final output of the teacher and student networks are identical. In the context of instance-level detection, it is not trivial to apply KD to high-res teacher and low-res student networks because they do not share the same feature/output spatial size at the same network stages, as illustrated in Fig.~\ref{fig:sketch}(a). Downsampling feature maps and output of the pre-trained large-resolution teacher to match those of the low-resolution student is one naive workaround. But this operation significantly corrupts predicted features and output, making them poorly reflective of the actual knowledge learned by the teacher.

We, instead, explore alignment of feature maps to resolve the output size mismatch between high-res teacher and low-res student. For the feature pyramid (FPN) structure~\cite{DBLP:conf/cvpr/LinDGHHB17} widely used in instance-level detection networks, the feature map size in last network stage is $2\times$ larger than that of current stage. Based on this observation, for the low-res student, we adopt input resolution $2\times$ smaller than the typical input used for the teacher. This provides feature-level consistency between the two input resolution models and allows their features to match spatially. As shown in Fig.~\ref{fig:sketch}(b), the spatial size of $P_2$ with low-res (downsampled by $2\times$) input shares the same spatial size as $P_3$ of the high-res input. This simple strategy quickly and effectively enables knowledge distillation from teacher to student. 

With this novel alignment idea, we propose an aligned multi-scale training method and a crossing feature-level fusion module to train a strong teacher. Aligned multi-scale training qualifies a ready-for-distillation robust teacher that performs well across multiple input resolutions. Whereas the crossing feature-level fusion module dynamically fuses the features from multiple-res models within the same teacher network. Finally, the rich multi-scale and multi-resolution knowledge of the \textit{multi-scale fusion teacher} is distilled to the low-res student, resulting in a high-performing low-resolution model. Fig.~\ref{fig:sketch}(b) provides a high-level overview of our approach. 

Our main contribution is threefold.
\begin{itemize}
    \setlength{\itemsep}{0pt}
    \item The alignment concept to align feature maps of models at different input resolutions.
    \item A framework for training a strong multi-scale and multi-resolution fusion teacher that provides more informative training signals to the low-res student that does not have access to fine visual details in high-res images.
    \item Extensive ablation studies and comparative experiments on different instance-level detection tasks to demonstrate the effectiveness of our methods.
\end{itemize}
        
\section{Related Work}
\subsection{Instance-level Detection Tasks}
Instance-level detection tasks, including object detection~\cite{girshick2014rich,dollar2015fast,DBLP:conf/nips/RenHGS15,DBLP:conf/cvpr/LinDGHHB17,lin2017focal,tian2019fcos,DBLP:conf/nips/DaiLHS16}, instance segmentation~\cite{he2017mask,DBLP:journals/corr/abs-1803-01534,xie2020polarmask,chen2019tensormask,zhang2020MEInst,qi2020pointins}, and key point detection~\cite{he2017mask,zhou2019objects,sun2019deep,xiao2018simple}, require detecting objects at the instance level. From the viewpoint of coarse-to-fine recognition, an instance can be represented by a bounding box in object detection, a pixel-wise mask in instance segmentation, and a sequence of key points.
 
Recently, single-shot instance-level detection methods~\cite{lin2017focal,liu2016ssd,redmon2016you,redmon2018yolov3,wang2019solo,ying2019embedmask,zhang2020MEInst,qi2020pointins,xie2020polarmask,tian2020conditional,qi2021open,li2021fully} have gained interest. Single-shot methods are aimed at accelerating model inference time while maintaining good detection performance, by designing new network modules and architectures. While network design is an important factor in determining runtime efficiency~\cite{polino2018model,chin2020towards,li2019exploiting,kim2018paraphrasing,ding2019global,zhuang2018discrimination,gao2018dynamic,zhu2017prune,carreira2018learning}, it is not the only way. In this paper, we take a new direction to improve runtime efficiency through input resolution reduction, without structurally modifying the network significantly.

\subsection{Knowledge Distillation}
\cite{hinton2015distilling} is a seminal work in knowledge distillation, which can be used to train a compact network by distilling knowledge from a larger teacher network. Over the years, many improved KD methods have been proposed that perform distillation over spatial attention \cite{zagoruyko2016paying}, intermediate features \cite{romero2014fitnets, heo2019comprehensive, tian2019contrastive}, relational representation \cite{park2019relational,tung2019similarity}, improved teachers \cite{cho2019efficacy, mirzadeh2020improved}, etc. In contrast to existing KD methods that focus on improving the performance of compact networks, the focus of this paper is on improving the low-resolution model with KD which poses unique challenges.

\subsection{Improving Low-Resolution Models}
Multi-scale training is commonly used in image classification \cite{kuen2018stochastic,mudrakarta2019k,wang2020resolution,yang2020mutualnet,li2020learning} and object detection \cite{he2014spatial, singh2018analysis, singh2018sniper,chen2021scale} to improve model robustness against input resolution variation. It is an easy approach to improve performance at multiple or even low input resolutions. Conventional multi-scale training does not involve KD and thus generally does not guarantee spatially-aligned features required for seamless KD in instance-level detection. There is work to apply KD to improve low-resolution image classification in conjunction with multi-scale training \cite{mudrakarta2019k,wang2020resolution,yang2020mutualnet,li2020learning}. This strategy is straightforward for image classification given that the multi-resolution models share the same output size. However, it is not the case for instance-level detection and there is difficulty to apply KD. To circumvent this difficulty, we propose an aligned multi-scale training method.

\begin{figure*}[t!]
	\begin{center}
		\includegraphics[width=1.0\linewidth]{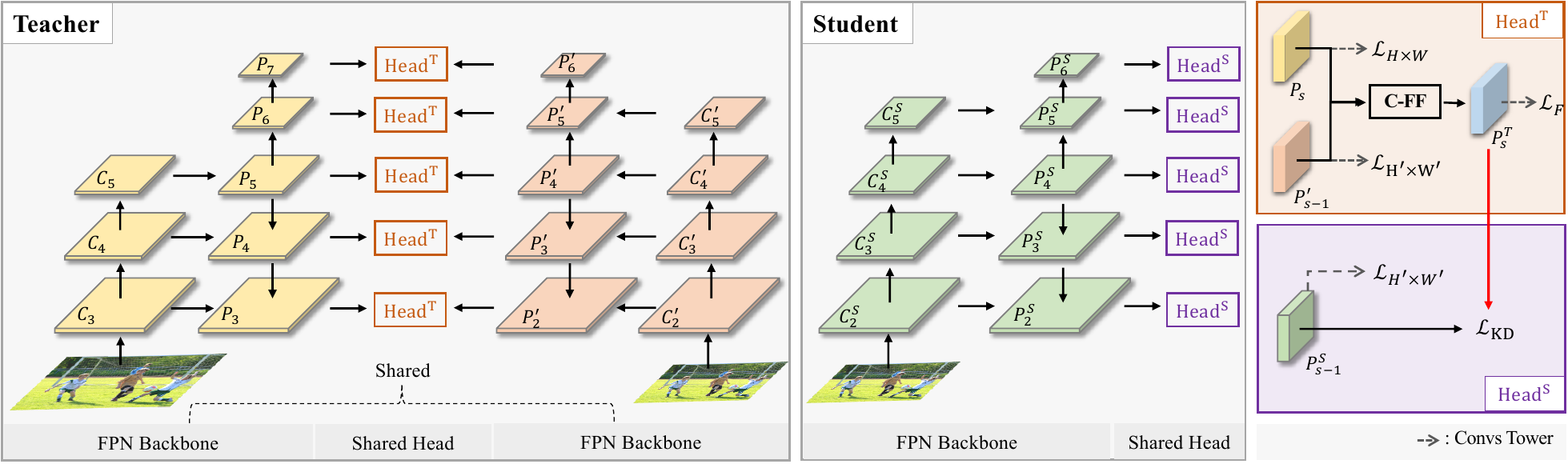}
	\end{center}
	\vspace{-13pt}
	\caption{Overview of the proposed multi-scale aligned distillation framework. $k$=$2$ is used here for illustration. ``Convs Tower'' refers to the convolution blocks in the detection head \cite{tian2019fcos}. In the first stage, we train a multi-scale teacher ($T$) that uses the same FPN~\cite{DBLP:conf/cvpr/LinDGHHB17} backbone for both high- and low-resolution input in an \textbf{aligned multi-scale training} fashion. The pyramid features from the two input resolutions are dynamically fused using the {crossing feature-level fusion} (C-FF) module. In the second stage, the trained multi-scale fusion teacher guides the low-resolution student ($S$) training via the distillation loss $\mathcal{L}_{\text{KD}}$.
	}
	\vspace{-0.1in}
	\label{fig:framework}
\end{figure*}

\section{Methodology}
Fig.~\ref{fig:framework} provides an overview of the proposed framework. 
The framework is divided into two stages where the first (left) trains a strong multi-resolution teacher, and the second stage (right) trains a low-resolution one with the guidance of the multi-resolution teacher.
In the following, we revisit a base detection method. For convenience, we adopt object detection as the base task to demonstrate our method, and take a strong one-stage detector FCOS~\cite{tian2019fcos} as the base detector. Our methods are applicable to other instance-level detection tasks. Then, we introduce the teacher formation process that involves our proposed aligned multi-scale training and crossing feature-level fusion. Finally, with a strong multi-resolution teacher, we propose crossing feature-level knowledge distillation to guide the training of low-resolution student effectively. 

\subsection{Base Detection Architecture}
As shown in the left of Fig~\ref{fig:framework}, our framework is based on FCOS~\cite{tian2019fcos}, in which FPN~\cite{DBLP:conf/cvpr/LinDGHHB17} backbone and a detection head perform pixel-level box prediction and regression to achieve object detection.

FPN backbone adopts a feature pyramid scheme to compute features at multiple scales for detecting objects at different sizes. Specifically, FPN backbone extracts feature maps $P_s \in \mathbb{R}^{H^{P_s} \times W^{P_s} \times 256}$ of several resolutions at different FPN levels from the input image $I \in \mathbb{R}^{H \times W \times 3}$, where $H$ and $W$ denote the height and width of the image respectively. $H^{P_s}$, $W^{P_s}$ refer to height and width of FPN feature maps, where $s \in \{2, 3, 4, 5, 6, 7\}$ indexes the \textit{level} of the multi-scale feature maps generated by FPN. The FPN feature maps are spatially smaller than the input image by factors of $\{4,8,16,32,64,128\}$.

The detection head has two output branches: classification branch and regression branch. Each has four convolutional blocks with convolutional layers and rectified linear unit (ReLU) layers. These blocks are shared among all FPN levels from $P_2$ to $P_7$. FPN features go through the detection head to perform instance-level classification and box regression. It trains with the following loss for a high-resolution input image with height $H$ and width $W$:
\begin{equation}
\mathcal{L}_{H\times W} = \mathcal{L}_{\text{cls}} + \mathcal{L}_{\text{reg}} + \mathcal{L}_{\text{ctr}},
\label{eq:L_HW}
\end{equation}
\noindent where $\mathcal{L}_{\text{cls}}$ is the classification loss, $\mathcal{L}_{\text{reg}}$ is the bounding box regression loss, and $\mathcal{L}_{\text{ctr}}$ is the centerness loss.

\subsection{Multi-Scale Fusion Teacher}\label{sec:teacher}
One of the most important factors for knowledge distillation is a strong teacher. Here, we focus on feature-level knowledge distillation that distills through the FPN's pyramidal features. In this section, we propose methods to train a teacher to distill strong multi-resolution knowledge at feature level \cite{romero2014fitnets}, to guide the training of a low-resolution student.

Note we do not consider final output-level (\eg, classification, regression outputs) knowledge distillation in this paper.
The output-level knowledge distillation works well for image classification tasks. But it is not straightforward to apply output-level distillation to instance-level detection due to the extreme imbalance of background and foreground classes. Further, feature-level knowledge distillation is more generally applicable than output-level distillation where the latter may involve different loss functions for various instance-level tasks.

Although the most straightforward approach is to train a single-scale high-resolution teacher to guide the student, the single-scale teacher is not aware of multi-resolution input. Thus its learned features may be poorly compatible with those of low-resolution student. To address this problem, we adopt and extend the widely-used multi-scale training strategy to train a strong multi-scale teacher, via feature pyramid alignment. 

It is common knowledge that multi-scale training only improves the teacher network at the network parameter/weight level and does not explicitly incorporate multi-scale information to the features of an input image at a given single resolution. To distill knowledge with enhanced multi-scale information to a low-resolution student, we introduce crossing feature-level fusion to dynamically fuse two-resolution features generated by the same teacher network on two input resolutions.

\vspace{-0.1in}
\paragraph{Aligned Multi-Scale Training.}
Multi-scale training perturbs the base input resolution ($H, W$) by rescaling it with a random scaling factor $\hat{\alpha}$ sampled from the range of $[\alpha_{\text{min}}, \alpha_{\text{max}}]$ (\eg, [0.8, 1.0]) at every training iteration. It can be seen as training multiple models that act at different perturbed input resolutions and share the same network parameters/weights. Within the same network, the high-resolution models supposedly have strong knowledge that can be distilled to the low-resolution models. However, knowledge distillation is nontrivial due to the spatial size mismatch between the output and feature maps of models acting at different input resolutions.

In the FPN structure, the spatial sizes of any two adjacent pyramidal feature maps differ by $2\times$ along each spatial dimension. Motivated by this observation, we adopt two base resolutions $((H, W), (H', W'))$ respectively for high- and low-resolution models that share the same network weights, where $H'=H/k$, $W'=W/k$, and $k$ is any \textit{valid}\footnote{$k$ preserves the expected number of FPN levels in the model.} even number. Such a reduction factor allows us to shift the FPN's position at backbone network and obtain FPN pyramidal feature maps whose spatial sizes match those of high-resolution model. 

For clarity, we denote this shift offset as $m$, where $m=k/2$.
In FCOS, a low-resolution ($k$=2 and $m$=1) model outputs pyramidal feature maps at $\{P^{'}_2, P^{'}_3, P^{'}_4, P^{'}_5, P^{'}_6\}$ levels that have the same spatial sizes as the default $\{P_3, P_4, P_5, P_6, P_7\}$ levels of high-resolution model. During training, both high- and low-resolution models are trained simultaneously in the multi-scale training fashion with distinctly-sampled $\hat{\alpha}$.

Put differently, the low-resolution model uses earlier network blocks to generate pyramidal features in order to spatially match the pyramidal features of the high-resolution model. With this alignment approach, all models trained using the lower-resolution input (obtained by varying $k$) is \textit{aligned} with the high-resolution model in terms of the pyramidal feature map sizes. This approach critically eliminates the feature-size inconsistency between models of multiple input resolutions and is beneficial for crossing-resolution knowledge distillation. The proposed aligned multi-scale training loss is defined as
\begin{equation}
\mathcal{L}_{\text{Align}} = \mathcal{L}_{H\times W} + \mathcal{L}_{H'\times W'},
\label{eq:L_align}
\end{equation}
where $\mathcal{L}_{H\times W}$ and $\mathcal{L}_{H'\times W'}$ are the losses for default-/high-resolution and low-resolution input respectively. For simplicity, we include just one low-resolution ($H'\times W'$) model in the loss function. The aligned multi-scale training can be easily extended to include multiple low-resolution models. 

\vspace{-0.1in}
\paragraph{Cross Feature-level Fusion.}\label{sec:fusion}
\label{sec:C-FF}
\begin{figure}[t]
	\begin{center}
		\includegraphics[
		width=0.9\linewidth
		]{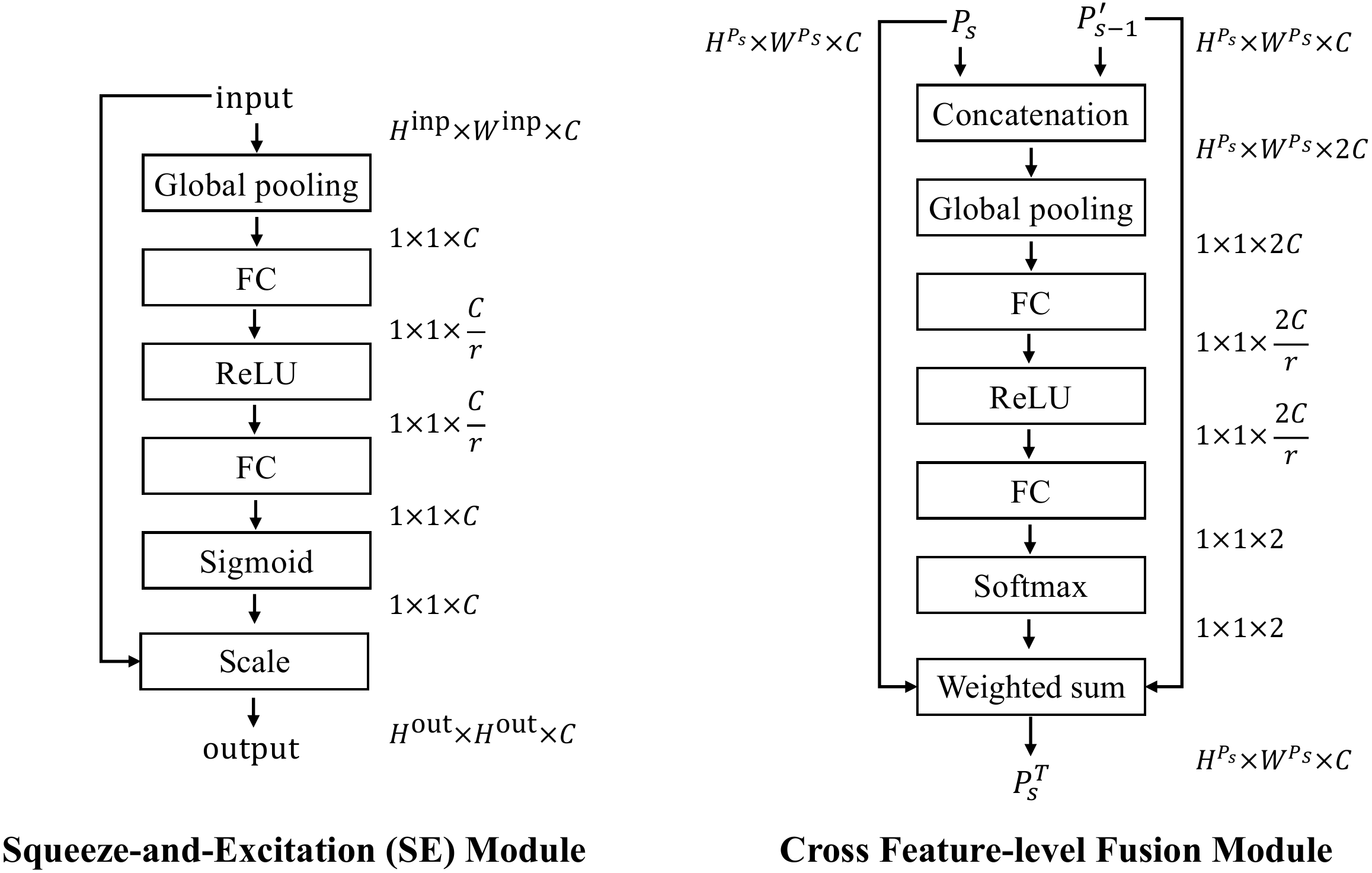}
	\end{center}
\vspace{-10pt}
	\caption{
	Comparison between 
	Squeeze-and-Excitation \cite{hu2018squeeze} module and our proposed crossing feature-level fusion module. In our module, $P_s$ and $P'_{s-1}$ refer to the
	features generated by
	high- and low-resolution input respectively.}
	\label{fig:ff}
\end{figure}

Intuitively, a high-resolution model performs better for small object detection due to better preservation of fine visual details in high-resolution images. Whereas, a low-resolution model works better for large object detection, as the backbone network captures information on the larger portions of the whole image, compared to high-resolution models at the same receptive field. This intuition is verified experimentally in Table~\ref{Tab:MS}. Although aligned multi-scale training encourages a network to be robust against multiple input resolutions, the network runs on just one of the ``seen" input resolutions during inference. Thus, its predicted features for any of the input resolutions do not incorporate the \textit{best of both worlds} from high- and low-resolution models.

Inspired by the Squeeze-and-Excitation (SE) module~\cite{hu2018squeeze}, we propose a feature fusion module to dynamically fuse pyramidal feature maps from different resolution input in an input-dependent manner.
Fig.~\ref{fig:ff} illustrates our feature fusion module.
It enables the network to adjust the degrees of contributions from different resolution input, depending on the content of the input image.

For example, an image with only large objects benefits more from features of low-resolution model, and vice versa. With this fusion module, the resulting features much improve.
Note that, in this paper, we only consider two-resolution input for easy demonstration of the idea. Actually, this module is readily extensible to fuse features in multiple resolutions. For each pair of feature maps, which share the same spatial sizes (\eg, $P_2$ and $P_3$ from low-resolution and high-resolution input with aligned multi-scale training), the fusion scores for each of the pair are dynamically predicted and used to fuse or combine them.

Initially, the module concatenates the two feature maps along the channel dimension and performs global average pooling to obtain 1D contextual features of
\begin{equation}
\begin{aligned}
P_s^p=&\frac{1}{H^{P_s}\times W^{P_s}}\sum_{i=1}^{H^{P_s}} \sum_{j=1}^{W^{P_s}} [P_s, P'_{s-m}](i,j),
\end{aligned}
\end{equation}
where $s\in \{3,4,5,6,7\}$ is for FCOS, $P$ and $P'$ are the pyramid feature maps of default-/high-resolution and low-resolution input images respectively. $H^{P_s}$ and $W^{P_s}$ are the height and width of the feature map $P_s$. $[\cdot]$ indicates the concatenate operation for two feature maps along the channel dimension. $P_s^p$ is then fed to a multi-layer perceptron (MLP) denoted as $\mathcal{H}$ to obtain Softmax-normalized fusion weights for the weighted sum that fuses\footnote{The same strategy is used to fuse two-/multi-resolution students with varying model complexity to counteract the loss of visual details needed for small object detection (see Table \ref{Tab:ME} and supplementary material).} the two feature maps of
\begin{equation}
P_s^T = h_s\{0\} \cdot P_s + h_s\{1\}\cdot P'_{s-m},
\end{equation}
where $h_s =  \mathcal{H}(P_s^p)$ and $\mathcal{H}$ is a MLP that comprises a FC layer with $\frac{2C}{r}$ output channels ($r$ is the channel compression ratio), ReLU function, a FC layer with 2 output channels, and Softmax function. $\{0\}$ and $\{1\}$ are the indexing mechanisms used to obtain the fusion scores for $P_s$ and $P'_{s-1}$ feature maps, respectively. In contrast to SE module that employs Sigmoid function and treats the output channels independently (\ie, multimodal distribution), we apply softmax normalization to explicitly encourage the ``either-or'' behavior (\ie, unimodal distribution) through the competitions among input of different resolutions.

Subsequently, the fused $P_s^T$ with strong multi-scale information is fed to the detection head for either training or inference. Given $P_s^T$, the training loss is defined as
\begin{equation}
\mathcal{L}_{F} = \lambda \cdot \mathcal{L}_{(H\&H')\times(W\&W')},
\label{eq:L_ff}
\end{equation}
where $(H\&H')\times (W\&W')$ indicates the use of the fused features from high- and low-resolution images. $\lambda$ is the loss weight.

\vspace{-0.1in}
\paragraph{Training Strategies.}
Since the fused multi-scale features are stronger than single-scale features from either high- or low-resolution input, we take the fused model as the strong multi-scale teacher to guide the training of low-resolution student in the next subsection. To obtain the fused multi-scale teacher, we train it with a two-step strategy, where the first stage performs aligned multi-scale training, and the second stage only trains the fusion module while ``freezing" the FPN and detection head. Alternatively, we can perform end-to-end training with joint aligned multi-scale training and feature fusion losses as
\begin{equation}
\mathcal{L}_{T} = \mathcal{L}_{\text{Align}} + \mathcal{L}_{F}.
\label{eq:L_ff}
\end{equation}
We empirically show that the two training strategies produce similar teacher's detection performance.

\subsection{Cross Feature-level Knowledge Distillation}\label{sec:distillation}
With aligned multi-scale training and crossing feature-level fusion, we obtain a strong multi-scale fusion teacher whose multi-resolution features can be seamlessly distilled to the low-resolution student. Similar to those in previous sections, we denote high resolution and input resolution as $H\times W$ and $H'\times W'$ used by the teacher and student respectively. Knowledge is distilled from teacher's features $P_s^T$ to student's $P^{S}_{s-m}$ via L1 loss as
\begin{equation}
\mathcal{L}_{\text{KD}} = \tau \cdot\sum_{s}{\lvert P_s^{T}-P_{s-m}^{S}\rvert},
\end{equation}
where $T$ and $S$ respectively refer to teacher and student, $s$ is the teacher's pyramidal feature level
(\eg, 3 to 7 for default input resolution in FCOS), $m$
is the shift offset used to spatially align student's feature maps with the teacher's, and $\tau$ is the loss weight hyperparameter. 
Following conventional knowledge distillation \cite{hinton2015distilling,romero2014fitnets}, the student is trained with both knowledge distillation loss and original detection loss, weighted by $\gamma$ as 
\begin{equation}
\mathcal{L}_{S} = \gamma \cdot \mathcal{L}_{\text{KD}} + (1 - \gamma) \cdot \mathcal{L}_{H' \times W'}.
\label{eq:L_ff}
\end{equation}

\section{Experiments}
\paragraph{Dataset \& Evaluation Metrics.}
We compare our method with state-of-the-art approaches on the challenging COCO dataset~\cite{lin2014microsoft}. Following common practice for COCO \cite{he2017mask,DBLP:conf/cvpr/LinDGHHB17,DBLP:journals/corr/abs-1803-01534,huang2019mask}, we use 115,000 training images and report evaluation results on the 5,000 validation images for the ablation experiments. 
Results on the 20,000 test-dev images are also reported for 
further comparison. We follow standard average precision 
metrics of AP (IoU range of $0.5$:$0.95$:$0.05$), $\text{AP}_{50}$ (IoU@0.5), $\text{AP}_{75}$ (IoU@0.75), $\text{AP}_{\mathbb{S}}$ (small-sized objects), $\text{AP}_{\mathbb{M}}$ (medium-sized objects), and $\text{AP}_{\mathbb{L}}$ (large-sized objects). 
To avoid confusion, we specify \textbf{AP$^{\bm{T}_\text{1}}$} for the teacher with only multi-scale training, \textbf{AP$^{\bm{T}_\text{2}}$} for the full multi-scale fusion teacher, and \textbf{AP$^{\bm{S}}$} for the student.


\vspace{-0.1in}
\paragraph{Implementation Details.}

All ablation experiments are conducted with FCOS with ResNet-50 \cite{he2016deep} backbone. For our final method, we perform evaluation on another popular detection method (RetinaNet \cite{lin2017focal}), and on other key instance-level tasks -- instance segmentation (Mask R-CNN \cite{he2017mask}), and keypoint detection (Mask R-CNN \cite{he2017mask}). Following existing practice in detection frameworks \cite{wu2019detectron2, chen2019mmdetection}, we train either teacher or student networks using batch size $16$ for $12$ epochs ($90,000$ iterations or $1\times$ schedule) in ablation experiments.

In general, we use high resolution $(800, 1,333)$ and set $(400, 677)$ as our {low resolution}. The first and second elements are image's short and maximum of long sides. For multi-scale training, $\alpha_{\text{min}}$ and $\alpha_{\text{max}}$ are set to 0.8 and 1.0 respectively. The two resolutions are denoted as \textbf{H} and \textbf{L}.

Stochastic gradient descent (SGD) with learning rate $0.01$ is used as the optimizer. We decay the learning rate with $0.1$ after $8$ and $11$ epochs for $1\times$ training schedule and scale these epoch numbers proportionally for longer training schedules. For teacher backbone networks, we initialize them with ImageNet pretrained models. Teacher and student models share the same training setting, except that we consider only single-scale low-resolution images for the student. 

By default, every low-resolution student shares the same backbone architecture and is initialized by its multi/high-resolution teacher. Note that we do not tune the hyperparameters for optimal performance. For all experiments (unless otherwise specified), we set $\lambda$, $\gamma$, and $\tau$ to 1.0, 0.2, and 3.0 respectively. The performance updates with respect to various hyperparameter values, as shown in Fig.~\ref{fig:merge_change}.

\subsection{Ablation Study}
Our proposed framework consists of two main stages:
multi-scale fusion teacher training and crossing feature-level knowledge distillation. We provide ablation study on them.


\subsubsection{Multi-Scale Fusion Teacher}

\begin{table}[t!]
    \centering
    \small
    \setlength{\tabcolsep}{2.5pt}
    \resizebox{\linewidth}{!}{%
        \small
        \begin{tabular}{c|c|ccc|ccc}
            \hline
             Training & Input & AP$^{T_1}$ & AP$_{50}^{T_1}$ & AP$_{75}^{T_1}$ & AP$_{\mathbb{S}}^{T_1}$ & AP$_{\mathbb{M}}^{T_1}$ & AP$_{\mathbb{L}}^{T_1}$ \\ 
             \hline
             \multirow{2}*{Single-scale} & H & 38.6 & 57.6 & 41.8 & 23.0 & 42.4 & 49.9\\ 
             & L & 34.1 & 51.6 & 36.1 & 15.0 & 37.6 & 50.8\\ 
             \hline
             \multirow{2}*{Vanilla} 
             \multirow{2}*{Multi-scale}
             & H & 40.3 & 59.3 & 43.8 & 25.6 &44.4 & 51.5 \\ 
             & L & 35.9 & 53.2 & 38.1 & 15.8 & 39.4 & 54.0\\ 
             \hline
             \multirow{1}*{Aligned} 
             \multirow{1}*{Multi-scale} 
             & H & 40.1 & 58.4 & 43.5 & 27.4 & 44.3 & 49.8 \\
             \multirow{1}*{(ours)}
             & L & 37.8 & 55.7 & 40.6 & 18.9 & 40.5 & 54.4 \\ 
             \hline
        \end{tabular}
    }
    \vspace{-7pt}
    \caption{Ablation study on multi-scale training for teacher.
    In single-scale training, $\hat{\alpha}$ is fixed to 1.0.
    $H$ and $L$ in column ``Input" indicates whether the inference is carried out with high- or low-resolution input.}
    \label{Tab:MS}
\end{table}

\vspace{-0.1in}
\paragraph{Aligned Multi-scale Training.}

The ablation study on multi-scale training is provided in Table~\ref{Tab:MS}. The teacher model trained with \textit{single-scale training} achieves $38.6$(H)/$34.1$(L) AP. With \textit{vanilla multi-scale training} using input of two base resolutions, the model obtains $40.3$(H)/$35.9$(L) AP.



The proposed aligned multi-scale training (with two base resolutions and feature map alignment) enables the teacher model to achieve better performance balance between H and L input resolutions at $40.1$(L)/$37.8$(H) AP. In particular, it improves over \textit{vanilla multi-scale training} by $1.9$ AP on low-resolution input. The gap between H and L is smaller than that of vanilla approach, indicating that the feature map alignment is important. It ensures that the model perform robustly against large variation of input resolutions.

\begin{figure*}[t!]
\centering
\includegraphics[width=\textwidth]{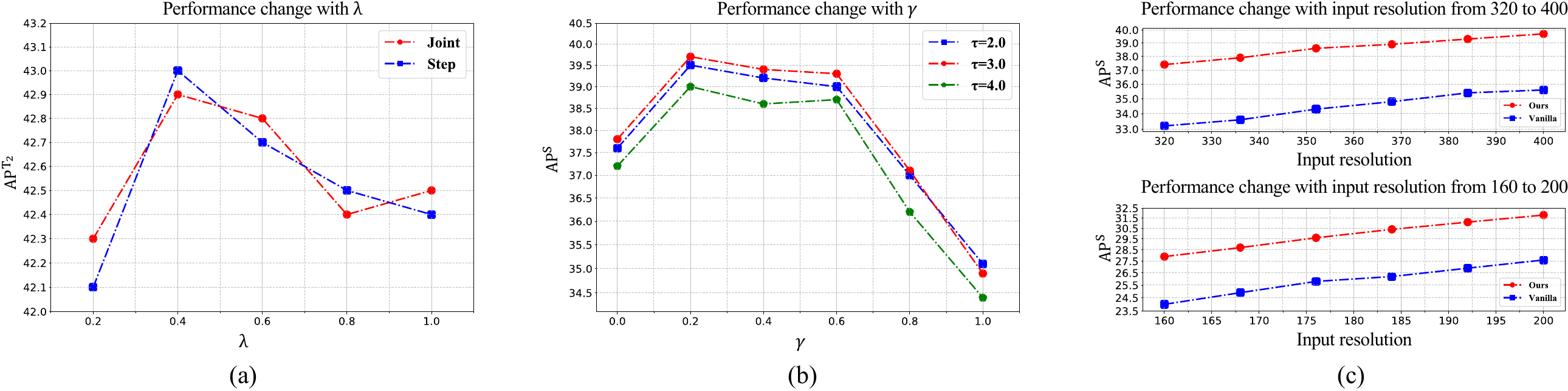}
\vspace{-10pt}\caption{
\textbf{(a)}: Performance change of teachers with respect to $\lambda$ and 
training strategy; \textbf{(b)}: Performance change of teachers with respect to $\gamma$ and $\tau$, \textbf{(c)}: Performance change of students with respect to inference resolution (320 to 400 \& 160 to 320).}
\label{fig:merge_change}
\end{figure*}

\begin{table*}[t!]\centering
\begin{minipage}{\textwidth}
        \begin{minipage}[t]{0.3\textwidth}
            \centering
            \small
            \setlength{\tabcolsep}{3pt}
            \begin{tabular}{c|ccc}
            \toprule
             \vspace{0.03in}
             Fusion approach & AP$^{T_2}$ & AP$_{50}^{T_2}$ & AP$_{75}^{T_2}$  \\ 
             \hline
             No fusion (H) & 40.1 & 58.4 &43.5 \\ 
             No fusion (L) & 37.8 & 55.7 &40.6 \\ \hline
             SC-SUM & 40.3 & 59.2 & 43.6  \\
             \hline
             CC & 41.4 & 60.3 & 44.7   \\ 
             \hline
             C-FF & 42.5 & 61.1 & 46.3 \\ 
             \hline
            \end{tabular}
            
        (a)
        \label{tab:C-FF-S1}
        \end{minipage}
    \begin{minipage}[t]{0.3\textwidth}
        \centering
        \small
        \setlength{\tabcolsep}{3pt}
        \begin{tabular}{c|ccc}
            \toprule
\vspace{0.03in}
             Design of C-FF & AP$^{T_2}$ & AP$_{50}^{T_2}$ & AP$_{75}^{T_2}$ \\ 
             \hline
             No fusion (H) & 40.1 & 58.4 &43.5 \\ 
             No fusion (L) & 37.8 & 55.7 &40.6 \\ \hline
             Sigmoid \cite{hu2018squeeze} & 41.6 & 59.7 & 45.0  \\ 
             \hline
             Channel-wise & 42.5 & 61.0 & 46.2  \\ 
             \hline
             C-FF & 42.5 & 61.1 & 46.3  \\ 
             \hline
        \end{tabular}
        
        (b)
    \end{minipage}
    \begin{minipage}[t]{0.35\textwidth}
        \centering
        \small
        \setlength{\tabcolsep}{2pt}
        \begin{tabular}{c|c|ccc}
            \toprule
\vspace{0.03in}
             Teacher & AP$^{T_2}$ & AP$^S$ & AP$_{50}^S$ & AP$_{75}^S$ \\ 
             \hline
             800 & 40.1 & 38.2 & 56.0 & 40.9 \\ \hline
             800\&400 ${\text{(CC)}}$ & 41.4 & 38.8 & 56.7 & 41.5 \\
             \hline
             800\&400 ${\text{(Sigmoid)}}$ & 41.6 & 38.9 & 56.6 & 41.8 \\
             \hline
             800\&400 ${\text{(C-FF)}}$ {$\lambda$=0.4} & 43.0 & 39.9 & 58.2 & 42.4  \\ 
             \hline
             800\&400 ${\text{(C-FF)}}$ {$\lambda$=1.0} & 42.5 & 39.7 & 58.0 & 42.5  \\ 
             \hline
        \end{tabular}
    
    (c)
    \end{minipage}
\vspace{-5pt}
\caption{Ablation studies. \textbf{(a)}: \textbf{Feature feature fusion approaches}: ``SC-SUM" sums the \textit{separately convolved} feature maps of two resolutions, ``CC" applies convolution to the concatenated features of the two resolutions, and ``C-FF" is our \textit{crossing feature-level fusion}. \textbf{(b)}: \textbf{Design of crossing feature-level fusion module}: ``Sigmoid" replaces the softmax with sigmoid activation as used in SE module \cite{hu2018squeeze}. ``Channel-wise" outputs channel-wise weights for feature fusion. \textbf{(c)}: \textbf{Effectiveness of different teachers} on guiding low-resolution student: single-scale 800px (high resolution) teacher and different multi-scale (800px \& 400px) fusion teachers with CC, sigmoid activation, or our crossing feature-level fusion.}
\label{Tab:part1}
\end{minipage}
\end{table*}

\begin{table*}[t!]
\begin{minipage}{\textwidth}
        \begin{minipage}[t]{0.4\textwidth}
            \centering
            \small
            \setlength{\tabcolsep}{2pt}
            \begin{tabular}{c|c|c|ccc}
            \toprule
\vspace{0.03in}
             Distillation & Type & Aligned  & AP$^S$ & AP$^S_{50}$ & AP$^S_{75}$ \\ 
             \hline
             \multirow{2}*{KD~\cite{hinton2015distilling}} & \multirow{2}*{Output} & $\circ$ & 36.4 & 54.0 & 38.6  \\
             & & $\checkmark$ & 37.5 & 55.8 & 39.7   \\ \hline
             \multirow{2}*{FGFI~\cite{wang2019distilling}} & \multirow{3}*{Feature} & $\circ$ & 36.6 & 54.3 & 38.8 \\ 
             & & $\checkmark$ & 39.5 & 57.9 & 42.2  \\ \cline{1-1} \cline{3-6}
            PODNet~\cite{douillard2020podnet} & & $\checkmark$ & 38.9 & 57.0 & 41.2  \\ \hline
            SKD~\cite{liu2020structured} & Attention & $\checkmark$ & 39.7 & 57.9 & 42.5 \\ \hline
            
             Ours  &  Feature & $\checkmark$ & 39.7 & 58.0 & 42.5  \\ 
             \hline
        \end{tabular}
        
        (a)
        \end{minipage}
    \begin{minipage}[t]{0.6\textwidth}
        \centering
        \small
        \setlength{\tabcolsep}{2pt}
        \begin{tabular}{c|c|c|ccc|ccc|c}
            \toprule
\vspace{0.03in}
             Input & Backbone  & 
             Width & AP$^S$ & AP$^S_{50}$ & AP$^S_{75}$ & AP$^S_{\mathbb{S}}$ & AP$^S_{\mathbb{M}}$ & AP$^S_{\mathbb{L}}$ & GFLOPS$_{\mathbf{b}}$ \\ \hline
             \multirow{6}*{L} & \multirow{4}*{R-50} &0.25$\times$ & 30.4 & 46.5 & 32.3 & 15.1 & 31.5 & 44.1 & 2.85 \\ \cline{3-10} & &0.50$\times$ & 36.1 & 53.7 & 38.8 & 18.4 & 38.2 & 50.4 & 8.56 \\ \cline{3-10} & &0.75$\times$  & 38.6 & 56.7 & 41.3 & 20.8 & 41.2 & 53.8 & 19.98 \\\cline{3-10} & & 1.00$\times$ & 39.7 & 58.0 & 42.5 & 21.7 & 42.9 & 55.0 & 36.03 \\ \cline{2-10}
             & R-101 & 1.00$\times$ &41.6 & 60.1 & 44.9 & 24.0 & 45.7 & 57.8 & 55.83 \\ \cline{2-10}
             & X-101 & 1.00$\times$& 43.1 & 61.7 & 46.5 & 25.7 & 47.3 & 59.4 & 74.57\\
             \hline
             H & R-50 & 1.00$\times$ & 42.3 & 61.3 & 45.9 & 28.2 & 45.9 & 53.5 & 142.69 \\  \hline
        \end{tabular}
        
        (b)
    \end{minipage}
    \label{Tab:part2}
\vspace{-7pt}
\caption{Ablation studies. \textbf{(a)}: \textbf{Distillation methods}. ``Type" indicates the type of knowledge distilled and ``Aligned" indicates whether the student's feature maps are aligned with the teacher's or not. All rows use the same multi-scale fusion teacher trained with C-FF. \textbf{(b)}: \textbf{Backbone architectures \& network widths}. ``GFLOPS$_b$" is the number of \textit{floating point operations} of the backbone network in giga unit. We average the GFLOPS computed over the 5,000 images per resolution type. The detection head is not included as it has a fixed GFLOPS of 57.09 regardless of the backbone architecture.}
\label{fig:part2}
\vspace{-0.1in}
\end{minipage}
\end{table*}

\vspace{-0.1in}
\paragraph{Cross Feature-level Fusion.}


Table~\ref{Tab:part1}(a) shows the effects of using different feature fusion strategies to fuse features of multi-resolution teacher models, compared to models without fusion. It reveals that models that employ feature fusion outperform single-resolution non-fusion ones, due to the incorporation of multi-resolution information in the fused features. Among the fusion models, our crossing feature-level fusion module that fuses features with dynamic fusion weights achieves better performance than both SC-SUM and CC that fuse features with static weights.

\begin{table*}[t!]
    \centering
    \small
    \centering
    \label{Tab:cc}
    \begin{tabular}{c|c|c|c|L{0.2cm}L{0.6cm}L{0.2cm}L{0.6cm}L{0.2cm}L{0.6cm}|L{0.2cm}L{0.6cm}L{0.2cm}L{0.6cm}L{0.2cm}L{0.8cm}}
        \hline
        Task & Detection Method & Training & Input & \multicolumn{2}{l}{AP} & \multicolumn{2}{l}{AP$_{50}$} & \multicolumn{2}{l|}{AP$_{75}$} & \multicolumn{2}{l}{AP$_{\mathbb{S}}$} & \multicolumn{2}{l}{AP$_{\mathbb{M}}$} & \multicolumn{2}{l}{AP$_{\mathbb{L}}$} \\ \hline
        
        \multirow{5}*{Object} 
        & \multirow{3}*{FCOS \cite{tian2019fcos}} & \multirow{2}*{Vanilla} & H & 42.8 & & 62.4 & & 46.7 && 27.3 && 46.5 && 55.2  \\ 
        \multirow{5}*{Detection} 
        & & & L & 38.2 && 56.3 && 41.0 && 17.8 && 42.2 && 56.7 \\ \cline{3-16}
        & & Ours & L & 41.6 & \textcolor[RGB]{34,139,34}{(+3.4)} & 59.9 &\textcolor[RGB]{34,139,34}{(+3.6)} & 44.9 & \textcolor[RGB]{34,139,34}{(+3.9)} & 22.8 &\textcolor[RGB]{34,139,34}{(+5.0)} & 44.8 & \textcolor[RGB]{34,139,34}{(+2.6)} & 57.0 & \textcolor[RGB]{34,139,34}{(+0.3)}\\ \cline{2-16}
        & \multirow{3}*{RetinaNet \cite{lin2017focal}} & \multirow{2}*{Vanilla} & H & 40.7 & & 60.7 & & 43.4 && 26.8 && 44.2 && 50.3 \\ 
        & & & L & 37.2 && 55.7 && 39.5 && 16.7 && 42.3 && 55.7  \\ \cline{3-16}
        & & Ours & L & 40.3 & \textcolor[RGB]{34,139,34}{(+3.1)} & 59.4 &\textcolor[RGB]{34,139,34}{(+3.7)} & 44.2 & \textcolor[RGB]{34,139,34}{(+4.7)} & 21.6 &\textcolor[RGB]{34,139,34}{(+4.9)} & 43.7 & \textcolor[RGB]{34,139,34}{(+1.4)} & 55.5 & \textcolor[RGB]{34,139,34}{(-0.2)} \\ \cline{1-16}
        
        \multirow{2}*{Instance}
        & \multirow{3}*{Mask R-CNN \cite{he2017mask}} & \multirow{2}*{Vanilla} & H & 39.8 && 57.6 && 42.9 && 19.8 && 44.5 && 57.8 \\
        \multirow{2}*{Segmentation}
        & & & L & 35.2 && 55.8 && 37.8 && 13.8 && 37.6 && 56.8\\ \cline{3-16}
        & & Ours & L & 37.3 & \textcolor[RGB]{34,139,34}{(+2.1)} & 58.2 &\textcolor[RGB]{34,139,34}{(+2.4)} & 39.7 & \textcolor[RGB]{34,139,34}{(+1.9)} & 16.5 &\textcolor[RGB]{34,139,34}{(+2.7)} & 39.4 & \textcolor[RGB]{34,139,34}{(+1.8)} & 57.3 & \textcolor[RGB]{34,139,34}{(+0.5)}\\ \cline{1-16}
        
        
        \multirow{2}*{Keypoint}
        & \multirow{3}*{Mask R-CNN \cite{he2017mask}} & \multirow{2}*{Vanilla} & H & 66.4 && 87.1 && 72.8 && 63.5 && 72.0 && 73.6 & \\
        \multirow{2}*{Detection}
        & & & L & 62.9 && 86.2 && 68.4 && 56.8 && 73.4 && 70.3  \\ \cline{3-16}
        & & Ours & L & 65.0 & \textcolor[RGB]{34,139,34}{(+2.1)} & 88.5 &\textcolor[RGB]{34,139,34}{(+2.3)} & 69.9 & \textcolor[RGB]{34,139,34}{(+1.5)} & 59.3 &\textcolor[RGB]{34,139,34}{(+2.5)} & 75.3 & \textcolor[RGB]{34,139,34}{(+1.9)} & 70.9 & \textcolor[RGB]{34,139,34}{(+0.5)} \\ \cline{1-16}
        
    \end{tabular}
    \caption{Overall performance evaluation on mainstream instance-level detection tasks with ResNet-50. ``Vanilla" refers to the standard multi-scale training. ``Ours" is the proposed framework with fusion teacher and crossing feature-level distillation. Here, we adopt $3\times$ training schedule to understand how our approach performs in the ``push-the-envelope" regime \cite{he2019rethinking}.}
    \label{Tab:instance}
\end{table*}

\begin{table}[t!]
    \centering
    \small
    \setlength{\tabcolsep}{3pt}
    \begin{tabular}{c|c|c|ccc|ccc}
        \hline
        Width & Role & Input & AP & AP$_{50}$ & AP$_{75}$ & AP$_{\mathbb{S}}$ & AP$_{\mathbb{M}}$ & AP$_{\mathbb{L}}$ \\ 
        \hline
        \multirow{4}*{1.0$\times$} 
        & \multirow{3}*{$T$} & H & 39.7 & 57.9 & 43.2 & 27.2 & 44.0 & 49.3 \\ \cline{3-9}
        & & L & 37.8 & 55.7 & 40.6 & 18.9 & 40.5 & 54.4 \\ \cline{3-9}
        & & H$\&$L & 42.5 & 61.1 & 46.3 & 28.0 & 45.7 & 55.7\\ \cline{2-9}
        & $S$ & L & 39.7 & 58.0 & 42.5 & 21.7 & 42.9 & 55.0 \\ \hline
        
        
        \multirow{4}*{0.5$\times$} 
        & \multirow{3}*{$T$} & H & 36.0 & 53.5 & 39.0 & 23.2 & 39.8 & 44.7\\ \cline{3-9}
        & & L & 33.8 & 50.8 & 36.4 & 16.3 & 35.4 & 49.1 \\ \cline{3-9}
        & & H\&L & 38.5 & 56.3 & 41.7 & 23.5 & 40.9 & 50.6\\ \cline{2-9}
        & $S$ & L & 36.1 & 53.7 & 38.8 & 18.4 & 38.2 & 50.4 \\ \hline
        
    \end{tabular}
    \vspace{-7pt}
    \caption{
    Performance evaluation on slimmed ResNet-50~\cite{yu2018slimmable} backbones. H\&L is the multi-scale fusion teacher that distills knowledge to student $S$.
    }
    \label{Tab:slimmable}
\end{table}

\begin{table}[t!]
    \centering
    \small
    \setlength{\tabcolsep}{3pt}
    \begin{tabular}{c|c|ccc}
        \hline
        Teacher  (H\&L) & Student  (L) & AP$^{S}$ & AP$^{S}_{50}$ & AP$^{S}_{75}$ \\ \hline
        R-50 & R-50 & 39.7 & 58.0 & 42.5\\ \hline
        \multirow{2}*{R-101} & R-50 & 40.5 & 58.8 & 43.7 \\ \cline{2-5}
        & R-101 & 41.6 & 60.1 & 44.9\\ \hline
    \end{tabular}
    \vspace{-7pt}
    \caption{Performance evaluation on multi-scale fusion teacher and low-resolution student models using distinct backbone architectures.}
    \label{Tab:tsdi}
\end{table}

\begin{table}[t]
    \centering
    \small
    \setlength{\tabcolsep}{2pt}
        \begin{tabular}{c|c|ccc|ccc}
            \hline
             Width (H) & Width (L)  & AP$^{S}$ & AP$^{S}_{50}$ & AP$^{S}_{75}$ & AP$^{S}_{\mathbb{S}}$ & AP$^{S}_{\mathbb{M}}$ & AP$^{S}_{\mathbb{L}}$ \\ \hline
             $0.50\times$ & $1.00\times$ & 41.4 & 59.9 & 44.5 & 24.6 & 44.5 & 54.6 \\ \hline
             $0.50\times$ & $0.75\times$ & 41.1 & 59.4 & 44.0 & 24.5 & 43.6 & 54.5 \\ \hline
             $0.50\times$ & $0.50\times$ & 40.1 & 58.5 & 43.2 & 24.5 & 42.7 & 51.8\\ \hline
             $0.25\times$ & $0.50\times$ & 37.6 & 55.5 & 40.6 & 20.8 & 40.0 & 50.9\\ \hline
        \end{tabular}
        \vspace{-7pt}
        \caption{Performance evaluation on using dual-resolution (high/H and low/L input resolutions) slimmed backbones within multi-scale fusion student models.}
        \label{Tab:ME}
\end{table}

Table~\ref{Tab:part1}(b) demonstrates the effects of changing output activation function and generating channel-wise output (\textit{i.e., $1\times1\times2C$}) in the crossing feature-level fusion (C-FF) module. Substituting the softmax activation in C-FF with Sigmoid (SE \cite{hu2018squeeze} module) degrades the AP by $0.9\%$. Softmax encourages the module to be more decisive when selecting features from either of the input resolutions, making the features not include unimportant resolutions. Generating channel-wise output to weight the two resolution's features does not bring any improvement.

\vspace{-0.1in}
\paragraph{Training Strategy.} 
Fig~\ref{fig:merge_change}(a) shows multi-scale fusion teacher's performance using different training strategies and $\lambda$. Two-step and joint training perform similarly. The best AP is achieved at $\lambda=0.4$.

\subsubsection{Cross Feature-level Knowledge Distillation}
\paragraph{Choice of Teacher.}
We analyze how different teacher variants affect low-resolution students' performance in Table~\ref{Tab:part1}(c). It shows that stronger teachers produce stronger students, and multi-scale fusion teachers produce the strongest students. We conjecture that multi-scale fusion teachers have greater compatibility with low-resolution students, due to the inclusion of low-resolution model's features in the feature fusion process.

\vspace{-0.1in}
\paragraph{Knowledge Distillation Methods.}
In Table~\ref{fig:part2}(a), we study the effect of using different distillation methods. Feature-level distillation methods work better than traditional output-level KD \cite{hinton2015distilling}, since intermediate features are more informative. With our alignment approach, all feature-level distillation methods perform quite similarly.

\vspace{-0.1in}
\paragraph{Loss Balancing Weights.}
Table~\ref{fig:merge_change}(b) shows the influence of the loss balancing weights $\tau$ and $\gamma$ 
on student's performance. $\gamma$ plays an important role in 
student's training, 
by balancing the knowledge distillation and original detection losses. $\tau$ can be easily tuned once $\gamma$ is fixed.

\vspace{-0.1in}
\paragraph{Inference at Multiple Resolutions.}
Table~\ref{fig:merge_change}(c) shows the AP performance of the student trained with three base resolutions, including $200$-pixel (short side) resolution with $P1$-$P5$ features via our alignment approach. Even at the small $200$-pixel resolution, our student achieves reasonably good AP $31.8$.

\vspace{-0.1in}
\paragraph{Different Backbones.}

We present the results of using different backbone architectures and network widths \cite{yu2018slimmable} to train teacher-student pairs using our framework in Table~\ref{fig:part2}(b). Here, teacher and student share the same backbone. Our multi-scale fusion teacher can also be used to train a strong high-resolution student (last row). Remarkably, our low-resolution ResNeXt-101 \cite{xie2017aggregated} (second last row) student outperforms the high-resolution R-50 student, at half of the required FLOPs. 

Our framework works very well with the slimmed/compact pre-trained backbones\footnote{Slimmable training  \cite{yu2018slimmable} is not used for detector training.
} \cite{yu2018slimmable} at multiple network widths. More results on slimmed backbones are reported in Table~\ref{Tab:slimmable}. Additionally, we report the results of using distinct teacher and student backbones in Table~\ref{Tab:tsdi}. Our approach benefits from using a stronger backbone for the teacher, like traditional KD \cite{hinton2015distilling}.

\vspace{-0.1in}
\paragraph{Multi-scale Fusion Students with Slimmed Backbones.}
In Table \ref{Tab:ME}, we show the performance of several such backbone combinations for the multi-scale fusion students that requires less computational footprints (FLOPS) than the full-width high-resolution model. Performance on small-sized objects $\text{AP}^{S}_{\mathbb{S}}$ is much improved compared to the models' single- and low-resolution counterparts reported in the supplementary material.



\subsection{Overall Performance Evaluation}
We apply our proposed framework to several recent methods across mainstream instance-level detection tasks of object detection, instance detection, and keypoint detection. The evaluation results are reported in Table~\ref{Tab:instance}. We notice that the low-resolution models trained with our approach outperform the widely-adopted \textit{vanilla multi-scale training} approach by $2.1\%$ to $3.6\%$, and performs competitively with its high-resolution models.

\section{Conclusion}
In this paper, we have boosted the performance on low-resolution instance-level detection tasks using our proposed framework. The framework comprises aligned multi-scale training and crossed feature-level fusion for training a strong teacher that dynamically fuses features from high-resolution and low-resolution input. By aligning the feature maps of teacher and student, knowledge of the multi-scale fusion teacher is correctly and effectively distilled to the low-resolution student. 

The extensive experiments demonstrate that our approach improves even strong baseline and vanilla multi-scale trained models by significant margins. Moreover, the proposed low-resolution detection approach is compatible with and complements compact networks (obtained with model compression techniques \cite{yu2018slimmable}) to reduce overall model complexity. We can extend our crossing feature-level fusion module to combine two lightweight models to achieve better instance-level detection performance, while maintaining low computational cost. 

\clearpage
{\small
\bibliographystyle{ieee_fullname}
\bibliography{egbib}
}
\end{document}